\documentclass[iicol]{sn-jnl}

\usepackage{graphicx}%
\usepackage{subcaption}
\usepackage{multirow}%
\usepackage{amsmath,amssymb,amsfonts}%
\usepackage{amsthm}%
\usepackage{mathrsfs}%
\usepackage[title]{appendix}%
\usepackage[dvipsnames]{xcolor}
\usepackage{textcomp}%
\usepackage{manyfoot}%
\usepackage{booktabs}%
\usepackage{algorithm}%
\usepackage{algorithmicx}%
\usepackage{algpseudocode}%
\usepackage{listings}%
\usepackage{adjustbox}
\usepackage{lmodern}
\usepackage{fix-cm}
\usepackage{hyperref}
\usepackage[capitalise]{cleveref}
\usepackage{pifont}

\usepackage{caption}
\usepackage{cite}
\usepackage[normalem]{ulem}

\usepackage{soul}

\begin{document}
\title[Article Title]{AEON: Adaptive Estimation of Instance-Dependent In-Distribution and Out-of-Distribution Label Noise for Robust Learning}


\author*[1]{\fnm{Arpit} \sur{Garg}}\email{arpit.garg@adelaide.edu.au}

\author[2]{\fnm{Cuong} \sur{Nguyen}}\email{c.nguyen@surrey.ac.uk}

\author[1]{\fnm{Rafael} \sur{Felix}}\email{rafael.felixalves@adelaide.edu.au}

\author[3]{\fnm{Yuyuan} \sur{Liu}}\email{yuyuan.liu@eng.ox.ac.uk}

\author[4]{\fnm{Thanh-Toan} \sur{Do}}\email{toan.do@monash.edu}

\author[1,2]{\fnm{Gustavo} \sur{Carneiro}}\email{g.carneiro@surrey.ac.uk}

\affil*[1]{\orgdiv{Australian Institute for Machine Learning}, \orgname{University of Adelaide}, \orgaddress{\country{Australia}}}

\affil[2]{\orgdiv{Centre for Vision, Speech and Signal Processing}, \orgname{University of Surrey}, \orgaddress{\country{United Kingdom}}}

\affil[3]{\orgdiv{Department of Engineering Science}, \orgname{University of Oxford}, \orgaddress{\country{United Kingdom}}}

\affil[4]{\orgdiv{Department of Data Science and AI}, \orgname{Monash University}, \orgaddress{\country{Australia}}}

\abstract{Robust training with noisy labels is a critical challenge in image classification, offering the potential to reduce reliance on costly clean-label datasets. Real-world datasets often contain a mix of in-distribution (ID) and out-of-distribution (OOD) instance-dependent label noise, a challenge that is rarely addressed simultaneously by existing methods and is further compounded by the lack of comprehensive benchmarking datasets. Furthermore, even though current noisy-label learning approaches attempt to find noisy-label samples during training, these methods do not aim to estimate ID and OOD noise rates to promote their effectiveness in the selection of such noisy-label samples, and they are often represented by inefficient multi-stage learning algorithms. 
We propose the \textbf{A}daptive \textbf{E}stimation of Instance-Dependent In-Distribution and \textbf{O}ut-of-Distribution Label \textbf{N}oise (\textbf{AEON}) approach to address these research gaps. AEON is an efficient one-stage noisy-label learning methodology that dynamically estimates instance-dependent ID and OOD label noise rates to enhance robustness to complex noise settings. Additionally, we introduce a new benchmark reflecting real-world ID and OOD noise scenarios. Experiments demonstrate that AEON achieves state-of-the-art performance on both synthetic and real-world datasets\footnote{Code will be open-sourced upon acceptance.}.}




\maketitle

\begin{figure}[t]
    \centering
\includegraphics[width=\linewidth]{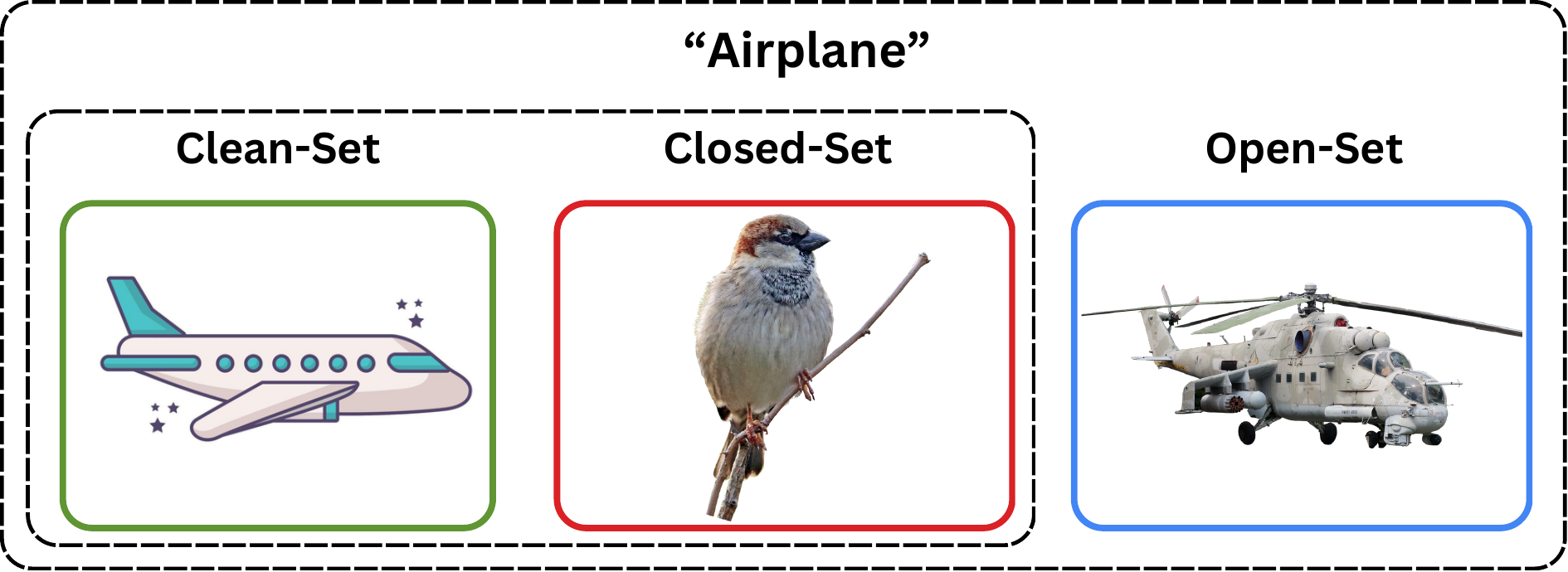}
    \caption{Different types of  samples labeled as ``Airplane'': Clean-Set ({\color{ForestGreen}$\bullet$}) has samples with correct labels, Closed-Set ({\color{BrickRed}$\bullet$}) contains samples with incorrect labels, where the image class (``Bird'') is in the set of training labels, and Open-Set ({\color{RoyalBlue}$\bullet$}) has samples with incorrect labels, where the image class (``Helicopter'') is not in the set of training labels. 
    }
    \label{fig:noise_types}
\end{figure}

\begin{figure*}[t]
    \centering
    \begin{subfigure}[b]{0.48\textwidth}
        \centering
        \includegraphics[width=\textwidth]{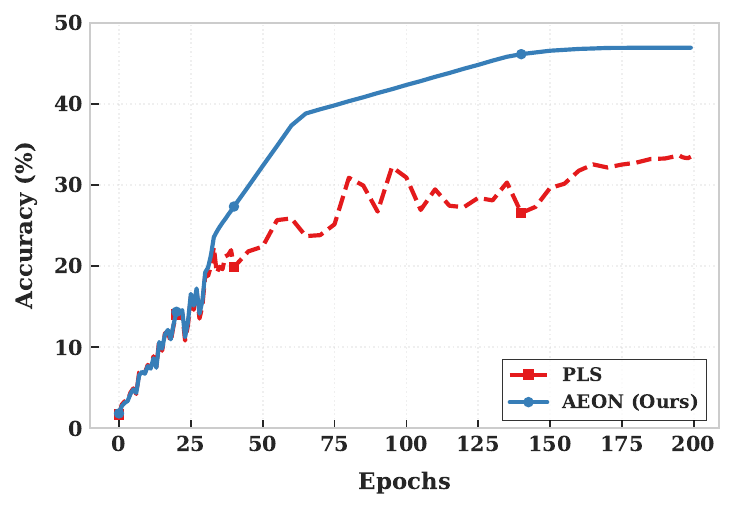}        \label{fig:accuracy}
    \end{subfigure}
    \hfill
    \begin{subfigure}[b]{0.48\textwidth}
        \centering
        \includegraphics[width=\textwidth]{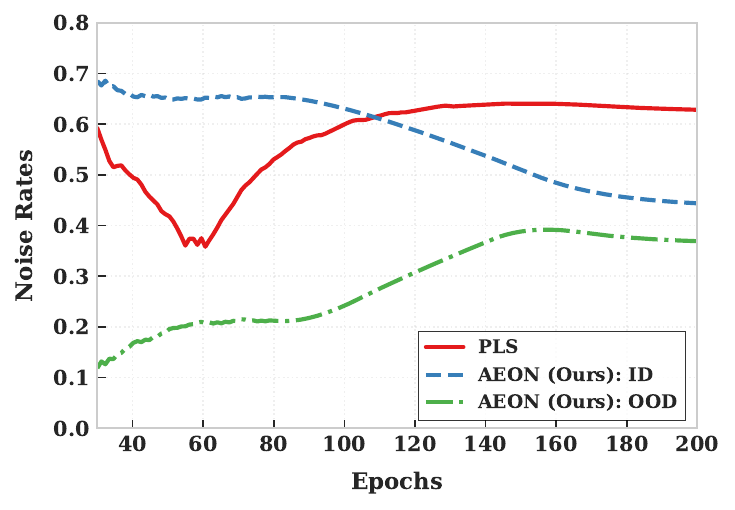}        \label{fig:noise_rates}
    \end{subfigure}
    \caption{
    Correlation between noise rate estimation (\(\hat{\eta}\)) and model performance on the ciFAIR-100 benchmark \cite{barz2020we} with \(40\%\) closed-set IDN (\(r^{in}\)) \cite{xia2020part} and \(40\%\) open-set noise (\(r^{out}\)) based on Places-IDN \cite{zhou2017places}. (Left) Classification accuracy of AEON compared to PLS~\cite{albert2023your} over training epochs. (Right) AEON’s  estimation of ID (\(\hat{\eta}^{in}\)) and OOD (\(\hat{\eta}^{out}\)) noise rates during training. For PLS, post-hoc noise estimation is shown using confidence values, as it does not directly estimate noise rates.}
   \label{fig:motivation}
\end{figure*}

\section{Introduction}\label{sec:intro}

Image classification has traditionally relied on meticulously curated datasets predominantly composed of clean-label samples~\cite{wu2020topological,zhang2024visionmamba}. However, real-world datasets often lack careful curation, resulting in significant proportion of instance-dependent label noise. This noise typically appears in two forms: (1) \textbf{closed-set or in-distribution (ID) noise}, where samples are incorrectly labelled but belong to the training categories, and (2) \textbf{open-set or out-of-distribution (OOD) noise}, where samples are mislabeled and originate from categories outside the training label set (\cref{fig:noise_types}).
Recent studies reveal that real-world datasets commonly exhibit ID and OOD instance-dependent label noise~\cite{northcutt2021pervasive,beyer2022better,li2017webvision,xiao2015learning}. Despite this, existing methods generally address these noise types independently with inefficient multi-stage learning algorithms. For ID noise, approaches often have one of the learning stages that use mixture models to classify samples into clean or noisy categories~\cite{arazo2019unsupervised,li2020dividemix}. For OOD noise, one of the learning stages usually employs energy-based scoring techniques to separate clean samples from noisy ones~\cite{albert2022embedding,liu2020energy}. 
This compartmentalized approach overlooks the coexistence of ID and OOD noise in real-world datasets, limiting the applicability of current methods in practical scenarios. Addressing this research gap remains a critical yet scarcely explored challenge in robust image classification.

Efforts to address both ID and OOD label noise simultaneously have emerged~\cite{wei2021open, sachdeva2021evidentialmix, albert2025accurate, albert2023your}, but these approaches often rely on ad-hoc parameters to reflect respective noise rates~\cite{li2020dividemix, han2018co, yao2020searching}. The accurate estimation of these noise rate parameters could significantly enhance their effectiveness, yet only few methods attempt to do so systematically. 
A notable example is our prior work~\cite{garg2024noiserate}, which estimates ID noise rates, but it does not try to jointly analyze ID and OOD label noise, leaving this critical gap largely unaddressed.
\cref{fig:motivation} demonstrates the value of estimating ID and OOD noise rates on our proposed benchmark (ciFAIR-100 with instance-dependent noise), containing both ID and OOD noise, where we observe that explicit noise rate estimation significantly enhances classification performance. Existing approaches like PLS~\cite{albert2023your} rely on confidence-based sample weighting without explicitly estimating noise rates during training, leading to suboptimal performance. In contrast,  our proposed method (AEON) achieves superior accuracy (\cref{fig:motivation}(a)) by precisely estimating and leveraging both ID and OOD noise rates in its training process (\cref{fig:motivation}(b)). 
Furthermore, most approaches treat OOD noise as random corruption~\cite{albert2025accurate, albert2023your, albert2022embedding}, ignoring its instance-dependent nature. This simplification overlooks the complexities of real-world datasets, where both ID and OOD noise are often instance-dependent~\cite{song2022learning, garg2024noiserate}. Compounding this issue is the prevalent use of synthetic benchmarks that, while incorporating both noise types, assume instance-independent OOD noise models~\cite{jiang2020beyond}. Such assumptions generate oversimplified scenarios~\cite{albert2022addressing}, which fail to capture the nuanced characteristics of realistic noise patterns~\cite{xia2020part, yao2021instance}.

This paper introduces the \textbf{A}daptive \textbf{E}stimation of Instance-Dependent In-Distribution and \textbf{O}ut-of-Distribution Label \textbf{N}oise for Robust Learning (AEON) method to address critical gaps in handling instance-dependent ID and OOD label noise. AEON is a novel approach for simultaneously estimating instance-dependent open-set and closed-set noise rates, enabling robust learning in scenarios with complex ID and OOD noise patterns.
We also introduce an efficient one-stage learning algorithm based on the AEON method that achieves competitive performance with only \(1.2\times\) computational overhead compared to the most efficient state-of-the-art (SOTA) approach~\cite{albert2023your}, making it practical for large-scale tasks.
Additionally, we propose a new ID+OOD instance-dependent label noise benchmark to capture the challenges of real-world datasets better. 
To summarize, our contributions are:
\begin{itemize}
    \item The novel AEON method, which jointly estimates open-set and closed-set noise rates to address instance-dependent ID and OOD label noise tasks effectively.  
    \item Efficient one-stage learning algorithm based on the AEON method that achieves competitive performance with only a \(1.2\times\) computational overhead compared to current SOTA approaches, making it feasible for large-scale applications.  
    \item ID+OOD instance-dependent label noise benchmark to systematically and comprehensively evaluate methods for learning with noisy labels in simulated instance-dependent ID and OOD noise scenarios.  
\end{itemize}
Our method achieves accurate ID and OOD noise rate estimation while maintaining computational efficiency comparable to existing SOTA methods. On standard benchmarks such as CIFAR-100~\cite{krizhevsky2009learning} with mixed noise (\(40\%\) closed-set and \(40\%\) open-set), our approach delivers a performance gain of approximately \(\approx 3\%\) in accuracy compared to SOTA approaches. More critically, on our challenging benchmark, we reach \(\approx 47\%\) accuracy, whereas competing methods fall below \(\approx 38\%\), highlighting the importance of our benchmark in challenging robust noisy-label learning methods.

\begin{figure*}[t]
    \centering
    \includegraphics[width=\textwidth,keepaspectratio]{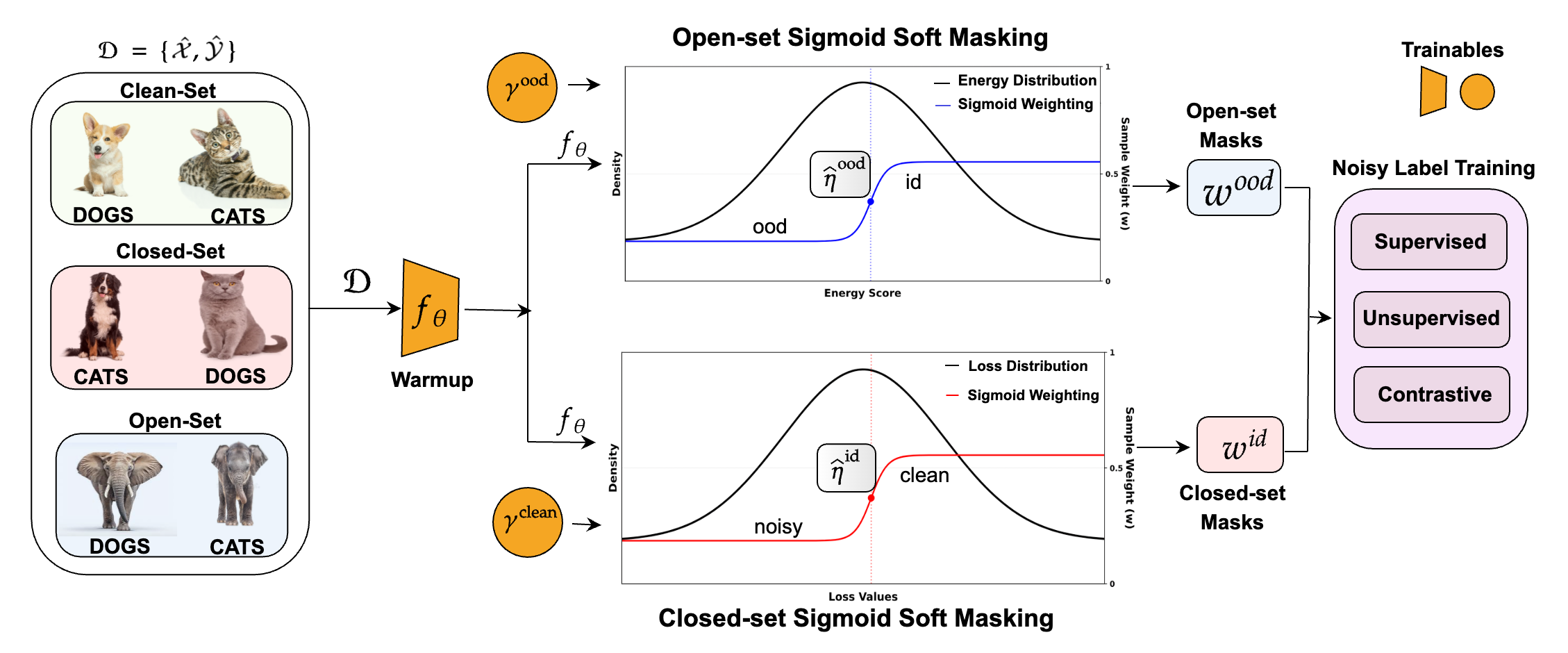}
    \caption{Our proposed  AEON is a novel end-to-end learning framework to simultaneously address instance-dependent closed-set and open-set label noise. 
    The framework comprises three key components: (1) a warm-up phase establishing initial feature representations through $f_\theta(.)$; (2) a dual-stream soft masking mechanism that dynamically estimates sample reliability through energy scores (for OOD label noise detection) and loss values (for ID label noise identification), producing adaptive weights $w^{ood}$ and $w^{id}$, respectively, via noise rates estimation ($\hat{\eta}^{id}$ for closed set, and $\hat{\eta}^{ood}$ for open-set); and (3) a unified multi-objective training strategy combining supervised learning on reliable ID samples, unsupervised learning on potentially noisy ID instances, and contrastive learning for robust feature discrimination on OOD samples. }
    \label{fig:framework}
\end{figure*}
\section{Related Work}\label{sec:related_work}

In this section, we first discuss the fundamental challenges of real-world label noise in \cref{subsec:real_world_noise},  showcasing the gap between theoretical models and practical applications. We then trace the evolution of noise detection methods in \cref{subsec:evolution_methods}, which reveals the limitations of traditional approaches in handling complex noise patterns. It naturally leads to our discussion of semi-supervised learning techniques in \cref{subsec:semi-supervised_methods}, which have emerged as a promising direction for leveraging noisy labels. We conclude with further exploration of instance-dependent noise modeling in \cref{subsec:idn_methods}, which expounds the more realistic scenario where noise patterns depend on sample instances.

\subsection{Real-world Label Noise}\label{subsec:real_world_noise}

The seminal work of \cite{angluin1988learning} laid the theoretical foundation for learning with label noise using ID instance-independent class-conditional transition matrices. While this model provided a strong theoretical basis, it did not address the complexities of real-world label noise. Empirical observations from real-world datasets~\cite{li2017webvision, wei2022learning} reveal that label noise often follows intricate patterns beyond simple instance-independent class-conditional transitions. Additionally, such datasets frequently include both ID and OOD label noise~\cite{albert2022addressing}, fundamentally challenging the closed-world assumptions that underpin earlier approaches to noise handling. Furthermore, many studies~\cite{xia2020part, liu2023generative, cheng2020learning} have established that label noise in real-world scenarios exhibits strong instance-dependent characteristics, where the likelihood of label corruption is instance-specific rather than class-dependent.

Synthetic benchmarks~\cite{krizhevsky2009learning,xia2020part, barz2020we} 
commonly used to evaluate noisy-label learning methods often fall short in accurately reflecting their performance on real-world datasets~\cite{xiao2015learning,li2017webvision, sun2021webly}.  
For instance, recent studies in webly-supervised learning have exposed critical limitations in existing noise-handling approaches~\cite{sun2021webly, wei2022learning, Hoang_2024_ACCV}, showing significant performance degradation on real-world datasets compared to synthetic benchmarks~\cite{sun2021webly, liu2022perturbed}. While subsequent work has advanced noise modeling by incorporating instance-dependent ID noise transitions and OOD samples~\cite{wei2021open}, existing synthetic benchmarks rely on instance-independent OOD samples that fail to capture the complexity of real-world noise patterns. 
To overcome these challenges, we present a new benchmark in \cref{subsec:benchmark_explained} that accurately reflects real-world noise distributions and provides a more rigorous evaluation setting for label noise methods.

\subsection{Evolution of Noise Detection Methods}\label{subsec:evolution_methods}
Early approaches to detect noisy-label samples exploited the observation that deep networks learn clean samples before memorizing noisy ones during training~\cite{arpit2017closer, liu2020early}. This observation gave rise to small-loss based selection methods, where samples with smaller loss values during training are deemed more likely to have clean labels. DivideMix~\cite{li2020dividemix} advanced this concept by modeling loss distributions using two Gaussian mixture models to separate clean and noisy samples. However, these approaches struggled with OOD samples, where the assumed bimodal distribution of losses no longer works effectively.
This issue has been addressed by some approaches, like EvidentialMix~\cite{sachdeva2021evidentialmix}, which jointly handle closed-set and open-set noise through evidential learning.

Recent advances in noisy-label sample detection mechanisms have demonstrated the power of feature-space analysis for handling noisy labels~\cite{kim2021fine,albert2022embedding,li2021learning}. For instance, methods such as SNCP~\cite{albert2022embedding} and RRL~\cite{li2021learning} achieve SOTA performance on real-world noisy datasets by effectively leveraging neighborhood relationships in learned representations.
Other relevant approaches explore feature component analysis to detect clean and noisy-label samples~\cite{kim2021fine}, where noisy label samples are detected using principal component analysis by reducing the feature space, visualizing the data for outliers, and identifying samples that deviate significantly from their class-specific clusters.

A common drawback of the methods discussed above is their reliance on inefficient two-stage processes, with one stage to identify clean and noisy-labeled samples, and another to train the model using separate strategies tailored to handle each type.
Moreover, existing approaches have not focused on a critical component that can lead to substantial improvement in their performance: the accurate noise rate estimation. 
Our proposed method, AEON, directly addresses this limitation while maintaining computational efficiency, advancing the field toward more robust learning under label noise.

\subsection{Semi-Supervised Learning with Noisy Labels}\label{subsec:semi-supervised_methods}

A significant breakthrough in noisy-label learning emerged by re-framing noise handling as a semi-supervised learning problem~\cite{liu2022acpl, liu2022perturbed, liu2025ittakestwo}, where the model treats potentially noisy-labeled samples as unlabeled data. DivideMix~\cite{li2020dividemix} pioneered this approach by leveraging noisy samples as unlabeled data, while subsequent methods like ScanMix~\cite{sachdeva2023scanmix} and PropMix~\cite{cordeiro2021propmix} enhanced this framework through semantic clustering and proportional mixing strategies, respectively. However, these approaches face two critical limitations: they rely on computationally expensive dual-model architectures, and their underlying distributional assumptions fail to address OOD samples. Our method overcomes both challenges through a single-model architecture that explicitly accounts for the ID and OOD label noise.

\subsection{Instance-Dependent Noise modeling}\label{subsec:idn_methods}
Recent theoretical breakthroughs have emphasized the importance of understanding and modeling instance-dependent noise (IDN) in real-world scenarios~\cite{cheng2022instance,xia2020part, liao2024instance, feng2024noisebox}, which established the learnability of instance-dependent noise under certain conditions. In contrast, \cite{berthon2021confidence} provided theoretical guarantees for learning with instance-dependent label noise through confidence scores. However, existing approaches rely on computationally expensive graphical models \cite{Garg_2023_WACV,garg2024noiserate},or make strong assumptions about the problem~\cite{albert2022addressing, yao2021instance}, or need to define complex thresholds about noise structure that limit their applicability~\cite{albert2023your,feng2021ssr}.

Even though successful in IDN problems, current SOTA IDN methods~\cite{garg2024noiserate} suffer from significant computational overhead, requiring \(3\times\) to \(5\times\) longer training time compared to standard noisy-label learning methods~\cite{Garg_2023_WACV, li2020dividemix, liu2022acpl}. Moreover, existing IDN approaches  focus either on closed-set or open-set noise~\cite{wang2023survey, cordeiro2021propmix, sachdeva2023scanmix}, failing to estimate crucial characteristics like noise rates for both scenarios jointly. This creates a fundamental bottleneck in learning, as accurate noise rate estimation is essential for robust model training~\cite{xia2020part, liu2023generative}. Our work addresses these limitations by introducing a computationally efficient framework that simultaneously estimates and handles both noise types while maintaining training costs comparable to standard procedures.





\section{Methodology}\label{sec:methodology}

In this section, we introduce the proposed AEON, a novel framework that effectively handles ID and OOD instance-dependent noise through adaptive noise rate estimation and robust learning. We first formalize the problem in \cref{subsec:problem_formulation}, then present our framework in \cref{subsec:framework}.

\subsection{Problem Formulation}\label{subsec:problem_formulation}
Let $\mathcal{\hat{X}} \subseteq \mathbb{R}^d$ denote the input space and $\mathcal{\hat{Y}} = \{1,\ldots,C\}$ the label space for $C$ classes. Given a noisy-label training dataset $\mathcal{D} = \{(\hat{x}_i, \hat{y}_i)\}_{i=1}^{|\mathcal{D}|}$, where $\hat{x}_i \in \hat{\mathcal{X}}$ represents potentially corrupted inputs and $\hat{y}_i \in \hat{\mathcal{Y}}$ their corresponding noisy labels, our goal is to learn a robust classifier $f_\theta: \mathcal{\hat{X}} \rightarrow \mathbb{R}^C$ parameterized by $\theta$. The labels $\hat{y}_i$ are $C$-dimensional one-hot encoded vectors from the set $\hat{\mathcal{Y}} = \{\hat{y} \in \{0,1\}^C | \mathbf{1}_C^{\top}\hat{y} = 1\}$, where $\mathbf{1}_C$ denotes a $C$-dimensional vector of ones. 

Taking a sample $(\hat{x},\hat{y}) \in \mathcal{D}$ and assuming that the latent clean label of $\hat{x}$ is $y$, we characterize two types of noise:
\begin{itemize}
    \item In-distribution (ID) noise rate $\eta^{id} = P(\hat{y} \ne y | \hat{x})$, where $y \in \hat{\mathcal{Y}}$
    \item Out-of-distribution (OOD) noise rate $\eta^{ood} = P(\hat{y} \ne y | \hat{x})$, where $y \notin \hat{\mathcal{Y}}$
\end{itemize}
Hence, a sample in $\mathcal{D}$ is annotated with a clean label with probability $1 - \eta^{id} - \eta^{ood} = P(\hat{y} = y | \hat{x})$.

The objective of our methodology is twofold: (1) estimate the noise rates $\eta^{id}$ and $\eta^{ood}$ without accessing a clean validation set, and (2) learn a model $f_{\theta}(.)$ that can accurately predict clean labels during testing.

\subsection{Noise Rate Estimation}\label{subsec:framework}

Our method estimates both ID and OOD noise rates through learnable parameters. We first explain how we estimate and detect OOD noise, followed by the ID noise estimation process.

\paragraph{OOD Noise Estimation}
We estimate the OOD noise rate, represented by $\hat{\eta}^{ood} \in [0,1]$, through a learnable parameter $\gamma^{ood} \in \mathbb{R}$ using a sigmoid function:
\begin{equation}\label{eq:eta_out}
\hat{\eta}^{ood} = \sigma(\gamma^{ood};T^{ood}) = \frac{1}{1 + e^{-\gamma^{ood}/T^{ood}}},
\end{equation}
where $T^{ood} \in \mathbb{R}^+$ controls the sigmoid smoothness. To detect OOD samples, we compute an energy score for each sample $i$:
\begin{equation}\label{eq:energy}
   E_i = -T_E \cdot \log \sum_{k=1}^C e^{f_{\theta}^{(k)}(\hat{x}_i)/T_E},
\end{equation}
where $f_{\theta}^{(k)}(\hat{x}_i) \in \mathbb{R}$ is the logit output for class $k$, $C$ is the number of classes, $\hat{x}_i$ is the input sample, and $T_E \in \mathbb{R}^+$ is a temperature parameter. We then compute a sample-wise OOD distribution weighting:
\begin{equation}\label{eq:w_dist}
   w^{ood}_i = \sigma(\tau^{ood} - E_i;\beta^{ood}) = \frac{1}{1 + e^{-(\tau^{ood}-E_i)/\beta^{ood}}},
\end{equation}
where $\beta^{ood} \in \mathbb{R}^+$ controls the sharpness of the sigmoid boundary, and $\tau^{ood}$ is an adaptive threshold:
\begin{equation}\label{eq:tau_E}
    \tau^{ood} = \Phi^{-1}(1 - \hat{\eta}^{ood}; \mu_E, \sigma^2_E),
\end{equation}
with $\mu_E$ and $\sigma^2_E$ being the empirical mean and variance of energy scores across the batch, and $\Phi^{-1}$ the inverse cumulative distribution function of the Gaussian distribution.

\paragraph{ID Noise Estimation}
Similarly, we estimate the ID noise rate $\hat{\eta}^{id} \in [0,1]$ through parameter $\gamma^{id} \in \mathbb{R}$:
\begin{equation}\label{eq:eta_in}
\hat{\eta}^{id} = \sigma(\gamma^{id};T^{id}) = \frac{1}{1 + e^{-\gamma^{id}/T^{id}}},
\end{equation}
where $T^{id} \in \mathbb{R}^+$ controls the sigmoid smoothness. For ID noise detection, we use the classification loss:
\begin{equation}\label{eq:supervised_loss}
   \mathcal{L}^s_i = -\hat{y}_i^{\top} \log \left(\mathsf{softmax}(f_{\theta}(\hat{x}_i))\right),
\end{equation}
where $\hat{y}_i \in \{0,1\}^C$ is the one-hot encoded label for sample $i$. We then compute a clean sample weighting:
\begin{equation}\label{eq:w_clean}
   w^{id}_i = \sigma(\tau^{id} - \mathcal{L}^s_i;\beta^{id}) = \frac{1}{1 + e^{-(\tau^{id}-\mathcal{L}^s_i)/\beta^{id}}},
\end{equation}
where $\beta^{id} \in \mathbb{R}^+$ controls the sigmoid boundary sharpness, and $\tau^{id}$ is an adaptive threshold:
\begin{equation}\label{eq:tau_L}
    \tau^{id} = \Phi^{-1}(1 - \hat{\eta}^{id}; \mu_L, \sigma^2_L),
\end{equation}
with $\mu_L$ and $\sigma^2_L$ being the empirical mean and variance of loss values.

These adaptive thresholds $\tau^{ood}$ and $\tau^{id}$ partition the energy and loss distributions according to the estimated noise rates. During training, if $\hat{\eta}^{ood} = 50\%$, then the threshold $\tau^{ood}$ will identify the 50\% higher energy samples as likely OOD. Similarly, $\tau^{id}$ uses the estimated $\hat{\eta}^{id}$ to identify noisy samples within the ID set through their loss values. 

\subsection{Training Algorithm}\label{subsec:learning}

We propose a one-stage learning algorithm with three complementary cost functions: (1) a supervised loss for clean-label samples, (2) an unsupervised loss for ID noisy-label samples, and (3) a contrastive loss for OOD noisy-label samples. A notable characteristic of our training is its end-to-end formulation, where samples are not explicitly assigned to one of the three objectives. Instead, their contributions are dynamically weighted using the estimated distribution and clean-label weights, $w^{ood}$ and $w^{id}$, as defined in Eqs.~\eqref{eq:w_dist} and \eqref{eq:w_clean}.

The overall optimization objective to learn $\theta^*, \gamma^{id^{*}}, \gamma^{ood^{*}}$ is defined as:
\begin{equation}\label{eq:training_obj}
\begin{split}
\min_{\theta,\gamma^{id},\gamma^{ood}} \frac{1}{N} \sum_{i=1}^{N} & \Big( w_i^{ood} \mathcal{L}_i^{id} + (1-w_i^{ood})\mathcal{L}_i^{ood} \Big) \\ 
&+ \mathcal{L}_{i}^{cont},
\end{split}
\end{equation}
where 
\begin{equation}\label{eq:training_loss}
\mathcal{L}_i^{id} = w_i^{id} \mathcal{L}_i^s + (1-w_i^{id}) \mathcal{L}_i^u + \max(0, E_i - m_{id})^2,
\end{equation}
is the loss for samples with high $w_i^{ood}$ (i.e., samples likely to belong to the training set classes), combining a supervised loss $\mathcal{L}_i^s$ for samples with high $w_i^{id}$, as defined in Eq.~\eqref{eq:supervised_loss}, and an unsupervised loss $\mathcal{L}_i^u$ for samples with low $w_i^{id}$
\begin{equation}\label{eq:unsup}
\mathcal{L}_i^u = -\hat{q}_i^{\top}\log p(f_\theta(\hat{x}_i)),
\end{equation}
with $p(f_\theta(\hat{x}_i))$ denoting the softmax probability output of the network, and the pseudo-label $\hat{q}_i$  being computed with
\begin{equation}\label{eq:unsup_normalise}
\hat{q}_i = \mathsf{normalize}\left(\left[\frac{1}{N}\sum_{n=1}^N p(f_\theta(\mathcal{T}_w^{(n)}(\hat{x}_i)))\right]^{\gamma_u}\right).
\end{equation}
In Eq.~\eqref{eq:unsup_normalise}, $\mathcal{T}_w^{(n)}$ denotes the $n$-th weak augmentation transformation, $N$ is the number of augmentations, and $\gamma_u$ controls pseudo-label sharpness, balancing prediction confidence with confirmation bias mitigation.
The loss for OOD samples in Eq.~\eqref{eq:training_obj} is defined as:
\begin{equation}\label{eq:loss}
\mathcal{L}_i^{ood} = \max(0, m_{ood} - E_i)^2,
\end{equation}
where $m_{out}$ and $m_{in}$ are margin hyperparameters controlling the energy bounds for OOD and ID samples respectively.

Finally, the contrastive loss $\mathcal{L}_i^{cont}$ in Eq.~\eqref{eq:training_obj} combines supervised and unsupervised objectives:
\begin{equation}\label{eq:cont}
\mathcal{L}_i^{cont} = \mathcal{L}_i^{cont,sup} + \mathcal{L}_i^{cont,uns},
\end{equation}
where the supervised contrastive loss is:
\begin{equation}\label{eq:cont_sup}
\mathcal{L}_i^{cont,sup} = -\log\frac{\sum_{j \in \mathcal{P}_i} \exp(\mathcal{S}_{ij})}{\sum_{k=1}^{N} \exp(\mathcal{S}_{ik})},
\end{equation}
and the unsupervised contrastive loss is:
\begin{equation}\label{eq:cont_unsup}
\mathcal{L}_i^{cont,uns} = -\log\frac{\exp(\mathcal{S}_{ii})}{\sum_{k=1}^{N} \exp(\mathcal{S}_{ik})}.
\end{equation}
Here, $\mathcal{S}_{ij}$ represents the similarity between feature representations of samples $i$ and $j$:
\begin{equation}\label{eq:unsup_similarity}
\mathcal{S}_{ij} = \frac{g(f_\theta(\mathcal{T}_w(\hat{x}_i)))^\top g(f_\theta(\mathcal{T}_s(\hat{x}_j)))}{T_c},
\end{equation}
where $g(.)$ is a projection head that maps features to a normalized embedding space, $\mathcal{T}_w$ and $\mathcal{T}_s$ are weak and strong augmentation functions respectively, $T_c$ is a temperature parameter, and $\mathcal{P}_i = \{j | \hat{y}_j = \hat{y}_i\}$ is the set of positive samples for instance $i$. The complete training procedure is summarized in Algorithm~\ref{alg:aeon}.

\section{Instance-Dependent Combined Open- and Closed-Set Noise Benchmark}\label{subsec:benchmark_explained}
We propose a systematic method to construct an instance-dependent ID and OOD label noise benchmark, utilizing the well-established ciFAIR-100 dataset~\cite{barz2020we}. This benchmark employs a two-stage noise injection strategy designed to mirror the complexities of real-world label noise, as detailed below.

\paragraph{Open-Set Noise Injection}
To introduce OOD noise, we first establish semantic relationships across datasets. Specifically, we compute the cosine similarity ($\Psi$) between feature representations of samples from ciFAIR-100~\cite{barz2020we}, denoted as $x^{\text{ciFAIR}}$, and Places365~\cite{zhou2017places}, denoted as $x^{\text{Places}}$, as follows:
\begin{equation}\label{eq:cosine_sim}
\Psi_{ij} = \frac{\psi(x_i^{\text{ciFAIR}})^\top \psi(x_j^{\text{Places}})}{\|\psi(x_i^{\text{ciFAIR}})\|_2 \|\psi(x_j^{\text{Places}})\|_2},
\end{equation}
where $\psi: \mathcal{X} \to \mathcal{Z}$ represents the mapping of input samples to the embedding space by the ImageNet pre-trained ResNet-18 encoder~\cite{he2016deep},  
and $\mathcal{Z}$ denotes the embedding space. 

Using these cosine similarity scores, we selectively replace a proportion of ciFAIR-100 training samples, determined by $r^{ood} \times |\mathcal{D}|$ (where $|\mathcal{D}|$ is the training set size), with semantically similar images from Places365. For each selected sample $\hat{x}_i$, we choose its replacement from Places365 by sampling from the top-$1$ most similar images according to $\Psi_{ij}$, while retaining the original ciFAIR-100 label $\hat{y}_i$. This creates instance-dependent OOD noise since both selection and replacement probabilities depend on feature similarities. We use Places365~\cite{zhou2017places} following previous OOD label noise papers~\cite{albert2022addressing, albert2022embedding, albert2023your, albert2025accurate, fooladgar2024manifold}. This process introduces instance-dependent OOD noisy-label samples that mirror real-world scenarios.

\paragraph{Closed-Set Noise Injection}
Subsequently, we inject instance-dependent closed-set noise on remaining clean samples following~\cite{xia2020part}. Specifically, for a noise rate $r^{id}$, we sample instance-specific flip rates $\mathbf{q} \in \mathbb{R}^d$ from $\mathcal{N}(\mu = r^{id}, \sigma = 0.1)$, clipping their values to be in the range $[0,1]$, 
and class-specific vectors $\mathbf{w}_c \in \mathbb{R}^d$ from $\mathcal{N}(\mu = 0, \sigma = 1)$. 
\begin{equation}\label{eq:noise_transition}
    P(\hat{y}|y=y_i, \hat{x}=\hat{x}_i) = \mathsf{softmax}(q_i \times  (\hat{x}_i \times \mathbf{w}_{y_i})),
\end{equation}
where the diagonal entry $P(\hat{y}=y_i|y=y_i, \hat{x}=\hat{x}_i)$ is set to $1-q_i$, and the off-diagonal entries are normalized to sum to $q_i$.  
This ensures that the instance-dependent noise transition depends on both feature characteristics and maintains an expected noise rate of $r^{id}$ across the dataset, refer~\cite{xia2020part} for more details.
\paragraph{Dual-Noise Framework}
For a dataset of size $|\mathcal{D}|$, we first replace $r^{ood} \times |\mathcal{D}|$ samples with instance-dependent OOD samples, followed by instance-dependent closed-set noise injection at rate $r^{id}$ on the remaining $(1-r^{ood}) \times |\mathcal{D}|$ samples. This dual-noise framework provides a rigorous testbed that effectively captures the coexistence of both noise types under instance-dependent conditions, closely resembling real-world data imperfections. Detailed experimental results on this benchmark are presented in~\cref{subsubsec:synthetic_benchmark}.


\section{Experiments}\label{sec:experiments}
In this section, we present comprehensive experiments across two categories of datasets to validate the effectiveness of our proposed method. First, we evaluate AEON on two synthetic benchmark datasets: (1) \emph{CIFAR-100}~\cite{krizhevsky2009learning} with random controlled noise injection for open-set (INet32~\cite{deng2009imagenet} and Places365~\cite{zhou2017places}) and random closed-set, examining multiple noise rate configurations following the literature~\cite{albert2025accurate,albert2023your,albert2022addressing}, and (2) \emph{ciFAIR-100}~\cite{barz2020we}, our novel ID+OOD instance-dependent synthetic benchmark incorporating realistic noise patterns from Places365~\cite{zhou2017places} for open-set and following IDN~\cite{xia2020part} for closed-set. Second, we assess AEON's performance on challenging \uline{real-world} datasets including \emph{Clothing1M}~\cite{xiao2015learning}, containing \(1M\) images from scrapping shopping websites; \emph{mini-WebVision}~\cite{li2020dividemix}, comprising \(66,000\) web-crawled images across \(50\) classes from WebVision~\cite{li2017webvision}; and the fine-grained \emph{WebFG-\(496\)}~\cite{sun2021webly} subsets (Web Aircraft, Web Bird, and Web Car) that present unique challenges for both open and closed-set noise. 
 
\begin{algorithm}[t]
\small
\caption{AEON: Unified Training Procedure}
\label{alg:aeon}
\begin{algorithmic}[1]
\Require $\mathcal{D}$, $f_\theta$, $(T_E, T^{id}, T^{ood}, T_c)$, $(m_{id}, m_{ood})$
\Ensure $f_\theta$, $(\hat{\eta}^{id}, \hat{\eta}^{ood})$
\State Init: $\theta$, $\gamma^{id}$, $\gamma^{ood}$

\While{not converged}
    \State {\footnotesize \textbf{1. Noise Rate Estimation}} \textbf{[Eq.~\eqref{eq:eta_out},\eqref{eq:eta_in}]:}
    \State $\hat{\eta}^{ood} \gets \sigma(\gamma^{ood};T^{ood})$ 
    \State $\hat{\eta}^{id} \gets \sigma(\gamma^{id};T^{id})$
    
    \State {\footnotesize \textbf{2. Detection Scores}} \textbf{[Eq.~\eqref{eq:energy},\eqref{eq:supervised_loss}]:}
    \State $E_i \gets -T_E\log\sum_{k=1}^C e^{f_\theta^{(k)}(\hat{x}_i)/T_E}$
    \State $\mathcal{L}^s_i \gets -\hat{y}_i^\top\log(\mathsf{softmax}(f_\theta(\hat{x}_i)))$
    
    \State {\footnotesize \textbf{3. Adaptive Thresholds}} \textbf{[Eq.~\eqref{eq:tau_E},\eqref{eq:tau_L}]:}
    \State $\tau^{ood} \gets \Phi^{-1}(1-\hat{\eta}^{ood}; \mu_E, \sigma^2_E)$
    \State $\tau^{id} \gets \Phi^{-1}(1-\hat{\eta}^{id}; \mu_L, \sigma^2_L)$
    
    \State {\footnotesize \textbf{4. Sample Weights}} \textbf{[Eq.~\eqref{eq:w_dist},\eqref{eq:w_clean}]:}
    \State $w^{ood}_i \gets (1 + e^{-(\tau^{ood}-E_i)/\beta^{ood}})^{-1}$
    \State $w^{id}_i \gets (1 + e^{-(\tau^{id}-\mathcal{L}^s_i)/\beta^{id}})^{-1}$
    
    \State {\footnotesize \textbf{5. Component Losses}} \textbf{[Eq.~\eqref{eq:unsup_normalise}-\eqref{eq:training_loss}]:}
    \State $\hat{q}_i \gets \mathsf{normalize}[\frac{1}{N}\sum_{n} p(f_\theta(T_w^{(n)}(\hat{x}_i)))]^{\gamma_u}$
    \State $\mathcal{L}^u_i \gets -\hat{q}_i^\top\log p(f_\theta(\hat{x}_i))$
    \State $\mathcal{L}^{ood}_i \gets \max(0, m_{ood} - E_i)^2$
    \State $\mathcal{L}^{energy}_i \gets \max(0, E_i - m_{id})^2$
    
    \State {\footnotesize \textbf{6. Contrastive Learning}} \textbf{[Eq.~\eqref{eq:unsup_similarity}-\eqref{eq:cont_unsup}]:}
    \State $\mathcal{S}_{ij} \gets g(f_\theta(T_w(\hat{x}_i)))^\top g(f_\theta(T_s(\hat{x}_j)))/T_c$
    \State $\mathcal{L}^{cont,sup}_i \gets -\log\frac{\sum_{j \in \mathcal{P}_i} \exp(\mathcal{S}_{ij})}{\sum_{k} \exp(\mathcal{S}_{ik})}$
    \State $\mathcal{L}^{cont,uns}_i \gets -\log\frac{\exp(\mathcal{S}_{ii})}{\sum_{k} \exp(\mathcal{S}_{ik})}$
    
    \State {\footnotesize \textbf{7. Final Loss \& Update}} \textbf{[Eq.~\eqref{eq:training_obj},\eqref{eq:training_loss}]:}
    \State $\mathcal{L}^{id}_i \gets w^{id}_i\mathcal{L}^s_i + (1-w^{id}_i)\mathcal{L}^u_i + \mathcal{L}^{energy}_i$
    \State $\mathcal{L}_{total} \gets \frac{1}{N}\sum_i [w^{ood}_i\mathcal{L}^{id}_i + (1-w^{ood}_i)\mathcal{L}^{ood}_i$ 
    \State $\quad\quad\quad\quad + \mathcal{L}^{cont,sup}_i + \mathcal{L}^{cont,uns}_i]$
    \State Update $\theta, \gamma^{id}, \gamma^{ood}$ using $\nabla\mathcal{L}_{total}$ 
\EndWhile
\State \Return $f_\theta, (\hat{\eta}^{id}, \hat{\eta}^{ood})$
\end{algorithmic}
\end{algorithm}

\subsection{Datasets}\label{subsec:datasets}
We conduct extensive experiments on both synthetic benchmark datasets in \cref{subsubsec:synthetic_benchmark} and real-world noisy datasets in \cref{subsubsec:real_benchmark} to demonstrate the effectiveness of AEON. Further details of datasets used in our experiments are outlined below.

\subsubsection{Synthetic Benchmarks}\label{subsubsec:synthetic_benchmark}
\textbf{CIFAR-100} comprises  CIFAR-100~\cite{krizhevsky2009learning} images $x \in \mathbb{R}^{32 \times 32 \times 3}$ across $N=100$ object categories, with $|\mathcal{D}_{\text{train}}| = 50,000$ and $|\mathcal{D}_{\text{test}}| = 10,000$. We simulate noise rates corruption following the literature~\cite{albert2022addressing, fooladgar2024manifold, albert2022embedding}. Following our algorithm to implement the benchmark explained in \cref{subsec:benchmark_explained}, we first corrupt a fraction $r^{id}$ of training labels while preserving images randomly for closed-set. Second, we replace $r^{ood}$ fraction of images from ImageNet32~\cite{deng2009imagenet} and Places365~\cite{zhou2017places}, maintaining consistent image dimensions while introducing domain shift for open-set.

\textbf{ciFAIR-100}, introduced in this paper, consists of ciFAIR-100~\cite{barz2020we} images $x \in \mathbb{R}^{32 \times 32 \times 3}$ across $N=100$ object categories, maintaining identical data distribution as CIFAR-100 with $|\mathcal{D}_{\text{train}}| = 50,000$ and $|\mathcal{D}_{\text{test}}| = 10,000$, but avoiding duplicates between train and test set~\cite{barz2020we}. We simulate open and closed noise rates corruption following the literature~\cite{albert2022addressing, fooladgar2024manifold, albert2022embedding}, using our benchmark construction algorithm explained in \cref{subsec:benchmark_explained}.

\subsubsection{Real-World Benchmarks}\label{subsubsec:real_benchmark}

\textbf{mini-WebVision} comprises web-crawled images $x \in \mathbb{R}^{227 \times 227 \times 3}$ from the first \(50\) classes of WebVision~\cite{li2017webvision}, with $|\mathcal{D}_{\text{train}}| = 66,000$ and $|\mathcal{D}_{\text{test}}| = 2,500$. Following DivideMix~\cite{li2020dividemix}, we evaluate on the clean set to ensure a fair comparison with prior work~\cite{li2020dividemix,garg2024noiserate,zhao2022centrality,zheltonozhskii2022contrast} on real-world noisy label learning.

\textbf{Clothing1M} consists of e-commerce web-crawled images~\cite{xiao2015learning} across $N=14$ clothing categories, with $|\mathcal{D}_{\text{noisy}}| = 1M$ web-collected training samples. The dataset has a clean-label test set with $|\mathcal{D}_{\text{test}}| = 10,000$ images. Following standard protocol~\cite{li2020dividemix, xia2020part}, we evaluate on the clean test set and have not considered the validation set at any stage.

\textbf{WebFG-496} contains fine-grained web images~\cite{sun2021webly} $x$ across $N=496$ categories, with $|\mathcal{D}_{\text{train}}| = 53,339$. The categories are partitioned into $|\hat{\mathcal{Y}}_{\text{bird}}| = 200$ bird species, $|\hat{\mathcal{Y}}_{\text{aircraft}}| = 100$ aircraft types, and $|\hat{\mathcal{Y}}_{\text{car}}| = 196$ car models. The dataset presents real-world challenges including natural label noise, minimal inter-class visual differences, and significant class imbalance, providing a rigorous benchmark for fine-grained recognition under noisy conditions.

\subsection{Implementation Details}\label{subsec:implementations}
In this section we provide implementation details of AEON, encompassing infrastructure setup, data processing, and optimization protocols. We ensure reproducibility by explicitly specifying all hyperparameters and architectural choices across different experimental settings.

\subsubsection{Infrastructure and Architecture}\label{subsubsec:infrastructure}
All experiments are conducted using PyTorch-\(2.5.1\) on NVIDIA \(A6000\) GPUs with FP32 precision and random seed 1 for reproducibility. We employ PreActResNet-18~\cite{he2016identity} architecture for CIFAR-100~\cite{krizhevsky2009learning}, ciFAIR-100~\cite{barz2020we}, and mini-WebVision~\cite{li2020dividemix}, pre-trained ResNet-50~\cite{he2016deep} for Clothing1M~\cite{xiao2015learning} and WebFG-496~\cite{sun2021webly} datasets following standard protocols in the literature~\cite{li2020dividemix,albert2025accurate,garg2024noiserate, yao2021jo}.

\subsubsection{Data Processing Pipeline}\label{subsubsec:processing}
Input images are processed at native resolutions: $32 \times 32 \times 3$ for CIFAR-100~\cite{krizhevsky2009learning} and ciFAIR-100~\cite{barz2020we}, $227 \times 227 \times 3$ for mini-WebVision~\cite{li2020dividemix}, and $448 \times 448 \times 3$ for WebFG-496~\cite{sun2021webly}, while Clothing1M~\cite{xiao2015learning} maintains original dimensions. For synthetic benchmarks (CIFAR-100~\cite{krizhevsky2009learning}, ciFAIR-100~\cite{barz2020we}), we augment inputs using random crop (padding=4) and horizontal flip for weak augmentation $\mathcal{T}_w$, while strong augmentation $\mathcal{T}_s$ applies RandAugment\((2,14)\) followed by CutOut\((16)\).

\subsubsection{Training and Optimization Protocol}\label{subsubsec:optimization}
For synthetic datasets, we optimize using SGD with momentum \(0.9\), initial learning rate \(0.1\), weight decay $5 \times 10^{-5}$, and batch size \(256\). Training proceeds for \(330\) epochs total, comprising \(30\) warm-up epochs followed by \(300\) main training epochs with cosine learning rate decay. Real-world datasets employ modified parameters: WebFG-496~\cite{sun2021webly} uses learning rate \(0.003\), weight decay $10^{-3}$, batch size \(32\), and \(10\) warm-up epochs, while Clothing1M~\cite{xiao2015learning} and mini-WebVision~\cite{li2020dividemix} follow similar settings as in~\cite{li2020dividemix, albert2023your}. These hyperparameters are selected based on performance on a held-out portion of the training set.

\paragraph{Noise Estimation Parameters}\label{subsubsec:noise_params}
The energy-based scoring mechanism, \cref{eq:energy} uses temperature $T_E=1$ with margins $m_{id}=0.2$ and $m_{ood}=0.8$ for synthetic datasets and $m_{id}=0.3$ and $m_{ood}=0.9$ for others. Adaptive rate estimation in \cref{eq:eta_out,eq:eta_in} employs temperature parameters $T^{ood} = T^{id} = 10$ with initial values $\gamma^{ood}$, $\gamma^{id}$ follow random initialization. Sample re-weighting in \cref{eq:w_dist,eq:w_clean} utilizes sigmoid temperatures $\beta^{id} = \beta^{ood} = 0.1$.

\paragraph{Multi-Objective Learning Configuration}\label{subsubsec:learning_config}
The supervised learning component in \cref{eq:supervised_loss} implements Mixup augmentation with $\alpha = 0.2$. Unsupervised learning in \cref{eq:unsup_normalise} uses temperature $\gamma_u=2$, and averages over $2$ weak augmentations. The contrastive learning module in \cref{eq:unsup_similarity} operates with temperature $T_c = 0.07$, projection dimension $128$, and balancing weight $\lambda = 0.5$ for all datasets.
Model selection follows established literature~\cite{li2020dividemix,albert2023your} for benchmark datasets. For synthetic and real-world experiments, we maintain consistent evaluation with recent literature~\cite{albert2022addressing, fooladgar2024manifold, albert2022embedding}.

\paragraph{Classification Measures}
We employ classification accuracy as our primary evaluation metric across all experiments. For synthetic benchmarks in \cref{tab:classification_comparison,tab:places_idn_results}, we additionally report Expected Calibration Error (ECE) to validate our model's uncertainty estimation. ECE evaluates the reliability of our dual-stream detection mechanism in \cref{eq:w_dist,eq:w_clean} and noise rate estimation in \cref{eq:eta_out,eq:eta_in} by quantifying how well the adaptive thresholds in \cref{eq:tau_E,eq:tau_L} partition the energy and loss distributions in \cref{fig:framework}. For real-world datasets in \cref{tab:clothing1m_webvision,tab:web_fine_grained_comparison}, we report accuracy to enable direct comparisons with prior work.

\subsection{Experimental Analysis}\label{subsec:comparisons_baseline}
We conduct extensive empirical evaluation of AEON across synthetic  and real-world benchmarks, analysing its performance against SOTA approaches. 

\paragraph{Performance on Our Instance-Dependent Open-set Noise Benchmark}
We establish new SOTA results on our proposed instance-dependent benchmark, as shown in \cref{tab:places_idn_results}. The evaluation encompasses recent approaches including DivideMix~\cite{li2020dividemix}, ELR~\cite{liu2020early}, EvidentialMix~\cite{sachdeva2021evidentialmix}, PLS~\cite{albert2023your}, and MDM~\cite{fooladgar2024manifold}. Under varying noise rate configurations, AEON achieves superior classification performance with margins ($\Delta_{acc}$) of $\{1.01\%, 4.03\%, 6.74\%, 9.15\%\}$ at $(r^{ood}, r^{id}) \in \{(0.2, 0.2), (0.4, 0.2), (0.6, 0.2), (0.4, 0.4)\}$ respectively. This improved performance is particularly significant given the challenging nature of instance-dependent noise that correlates with image content. The ECE improvements follow a similar pattern, with AEON reducing calibration error by $\{0.35\%, 1.09\%, 2.83\%, 3.57\%\}$ compared to the best baseline for the same noise configurations. These gains in calibration quality reinforce AEON's superior handling of instance-dependent noise.

\begin{table*}[t]
\centering
\caption{\textbf{Our benchmark}. Classification accuracy and ECE (\%) on ciFAIR-100~\cite{barz2020we} under instance-dependent open-set noise from Places365~\cite{zhou2017places} and instance-dependent closed-set noise from Part-Dependent~\cite{xia2020part}. The first two columns ($r^{ood}$, $r^{id}$) indicate injected open-set and closed-set noise rates respectively, while the last two columns show AEON's estimated noise rates ($\hat{\eta}^{ood}$, $\hat{\eta}^{id}$). All results are reproduced locally as the mean over three runs, experimental settings mentioned in \cref{subsec:implementations}, with \textbf{bold} indicating SOTA.}
\label{tab:places_idn_results}
\resizebox{\textwidth}{!}{%
\begin{tabular}{@{}lccccccccccc@{}}
\toprule
OOD & $r^{ood}$ & $r^{id}$ & CE & DM~\cite{li2020dividemix} & ELR~\cite{liu2020early} & EDM~\cite{sachdeva2021evidentialmix} & PLS~\cite{albert2023your} & MDM~\cite{fooladgar2024manifold} & \multicolumn{3}{c@{}}{\textbf{AEON (Ours)}} \\ 
& & & & \footnotesize{2020} & \footnotesize{2020} & \footnotesize{2021} & \footnotesize{2023} & \footnotesize{2024} & Acc./ECE & \multicolumn{2}{c}{Noise Rates} \\
\cmidrule(l){10-10} \cmidrule(l){11-12}
& & & & & & & & & ($\uparrow$/$\downarrow$) & $\hat{\eta}^{ood}$ & $\hat{\eta}^{id}$ \\ 
\midrule
\multirow{4}{*}{Places-IDN~\cite{zhou2017places}} 
& 0.2 & 0.2 & 48.07/7.82 & 73.10/5.94 & 65.27/6.45 & 68.94/6.12 & 73.50/2.89 & 70.28/3.45 & \textbf{74.51/3.24} & 0.24 & 0.31 \\
& 0.4 & 0.2 & 39.52/8.45 & 63.13/6.78 & 58.32/7.12 & 54.47/7.34 & 65.25/5.67 & 62.90/3.82 & \textbf{69.28/3.58} & 0.43 & 0.25 \\
& 0.6 & 0.2 & 27.58/9.67 & 41.67/7.89 & 42.77/7.56 & 40.65/8.23 & 42.43/3.34 & 44.63/4.15 & \textbf{51.37/3.92} & 0.57 & 0.23 \\
& 0.4 & 0.4 & 21.35/10.23 & 29.54/8.45 & 29.91/8.67 & 21.40/3.12 & 33.45/4.23 & 37.79/4.38 & \textbf{46.94/4.15} & 0.36 & 0.44 \\ 
\bottomrule
\end{tabular}}
\end{table*}

\paragraph{Performance on Synthetic Open-set Benchmarks}
\cref{tab:classification_comparison} presents comparative analysis on CIFAR-100~\cite{krizhevsky2009learning} with controlled open-set noise from INet32~\cite{deng2009imagenet} and Places365~\cite{zhou2017places}. AEON demonstrates consistent performance improvements across noise rate configurations $(r^{ood}, r^{id}) \in \{(0.2, 0.2), (0.4, 0.2), (0.6, 0.2), (0.4, 0.4)\}$, achieving average classification accuracy gains of $\Delta_{acc} = 2.25\%$ and $\Delta_{acc} = 3.1\%$ on INet32~\cite{deng2009imagenet} and Places365~\cite{zhou2017places}, respectively. Notably, the performance margin widens at higher noise rates (up to $3.47\%$ on INet32 and $4.12\%$ on Places365 at $(r^{ood}, r^{id})=(0.6, 0.2)$), suggesting enhanced robustness to severe corruption.  ECE analysis reveals consistent calibration improvements, with AEON reducing calibration error by an average of $2.15\%$ on INet32 and $2.85\%$ on Places365. The improvements are most pronounced at severe noise levels, reaching $3.27\%$ and $3.89\%$ reductions on INet32 and Places365 respectively at $(r^{ood}, r^{id})=(0.6, 0.2)$, aligning with the accuracy improvements.

\begin{table*}[t]
\centering
\caption{\textbf{Synthetic
Benchmarks}. Classification accuracy and ECE (\%) on CIFAR-100~\cite{krizhevsky2009learning} under random open-set from ImageNet32~\cite{deng2009imagenet} and Places365~\cite{zhou2017places}, and random closed-set noise~\cite{albert2023your}. The first two columns ($r^{ood}$, $r^{id}$) indicate injected open-set and closed-set noise rates respectively, while the last two columns show AEON's estimated noise rates ($\hat{\eta}^{ood}$, $\hat{\eta}^{id}$). Results are reported as the mean over three runs, with \textbf{bold} indicating SOTA. Baseline results from~\cite{albert2023your, fooladgar2024manifold, albert2025accurate}.}
\label{tab:classification_comparison}
\resizebox{\textwidth}{!}{%
\begin{tabular}{@{}lccccccccccccc@{}}
\toprule
OOD & $r^{ood}$ & $r^{id}$ & CE & ELR~\cite{liu2020early} & EDM~\cite{sachdeva2021evidentialmix} & RRL~\cite{li2021learning} & DSOS~\cite{albert2022addressing} & SNCF~\cite{albert2022embedding} & PLS~\cite{albert2023your} & MDM~\cite{fooladgar2024manifold} & \multicolumn{3}{c@{}}{\textbf{AEON (Ours)}} \\ 
& & & & \footnotesize 2020 & \footnotesize 2021 & \footnotesize 2021 & \footnotesize 2022 & \footnotesize 2022 & \footnotesize 2023 & \footnotesize 2024 & Acc./ECE & \multicolumn{2}{c}{Noise Rates} \\ 
\cmidrule(l){12-12} \cmidrule(l){13-14}
& & & & & & & & & & & ($\uparrow$/$\downarrow$) & $\hat{\eta}^{ood}$ & $\hat{\eta}^{id}$ \\
\midrule
\multirow{4}{*}{INet32~\cite{deng2009imagenet}} 
& 0.2 & 0.2 & 63.68 & 68.71 & 71.03 & 72.64 & 70.54 & 72.95 & \textbf{76.29} & 75.30 & 76.20/\scriptsize{3.21} & 0.23 & 0.26 \\
& 0.4 & 0.2 & 58.94 & 63.21 & 61.89 & 66.04 & 62.49 & 67.62 & 72.06 & 69.10 & \textbf{74.90/\scriptsize{3.45}} & 0.36 & 0.24 \\
& 0.6 & 0.2 & 46.02 & 44.79 & 21.88 & 26.76 & 49.98 & 53.26 & 57.78 & 59.80 & \textbf{63.27/\scriptsize{3.82}} & 0.63 & 0.23 \\
& 0.4 & 0.4 & 41.39 & 34.82 & 24.15 & 31.29 & 43.69 & 54.04 & 56.92 & 63.00 & \textbf{65.78/\scriptsize{3.94}} & 0.37 & 0.42 \\ 
\midrule
\multirow{4}{*}{Places365~\cite{zhou2017places}} 
& 0.2 & 0.2 & 59.88 & 68.58 & 70.46 & 72.62 & 69.72 & 71.25 & 76.35 & 74.50 & \textbf{77.89/\scriptsize{3.18}} & 0.22 & 0.29 \\
& 0.4 & 0.2 & 53.46 & 59.47 & 58.01 & 58.60 & 59.47 & 64.03 & 71.65 & 69.30 & \textbf{73.26/\scriptsize{3.52}} & 0.45 & 0.27 \\
& 0.6 & 0.2 & 39.55 & 37.10 & 23.95 & 49.27 & 35.48 & 49.83 & 57.31 & 59.00 & \textbf{60.74/\scriptsize{3.89}} & 0.58 & 0.29 \\
& 0.4 & 0.4 & 32.06 & 34.71 & 20.33 & 26.67 & 29.54 & 50.95 & 55.61 & \textbf{62.50} & 58.20/\scriptsize{4.12} & 0.43 & 0.36 \\ 
\bottomrule
\end{tabular}%
}
\end{table*}

\paragraph{Real-world Performance Analysis}\label{subsubsec:exp_real}
Our evaluation of large-scale real-world datasets demonstrates AEON's performance capability, where noise patterns emerge naturally from web collection processes. On Clothing1M~\cite{xiao2015learning}, AEON achieves $75.5\%$ Top-1 accuracy without using clean validation data during training, as shown in \cref{tab:clothing1m_webvision}. The estimated noise rate aligns with previous studies~\cite{xiao2015learning, garg2024noiserate}, validating our noise modeling approach. Similarly, on mini-WebVision~\cite{li2020dividemix}, we observe competitive performance with Top-1/Top-5 accuracies of $79.8\%/94.1\%$.

\begin{table}[t]
\centering
\caption{\textbf{Real-world Clothing1M and mini-WebVision Benchmarks.} Classification accuracy (\%) on Clothing1M~\cite{xiao2015learning} (\(N = 14\) clothing categories, \(|\mathcal{D}|=1M\)) and mini-WebVision~\cite{li2020dividemix} (\(N = 50\) object categories, \(|\mathcal{D}|=66K\)). Results report both Top-\(1\) and Top-\(5\) performance, with \textbf{bold} indicating SOTA.}
\label{tab:clothing1m_webvision}
\begin{tabular}{lccc}
\toprule
\multirow{2}{*}{Method} & Clothing1M & \multicolumn{2}{c}{mini-WebVision} \\ 
\cmidrule(lr){2-2} \cmidrule(lr){3-4}
& Accuracy ($\uparrow$) & Top-1 ($\uparrow$) & Top-5 ($\uparrow$) \\ \midrule
DivideMix~\cite{li2020dividemix}          & 74.6 & 77.2 & 91.6 \\
ELR+~\cite{liu2020early}                  & 71.5 & 63.6 & 83.5 \\
CC-GM~\cite{garg2024noiserate}             & 75.3 & \textbf{80.0} & 93.8 \\
MDM~\cite{fooladgar2024manifold}                                      & 73.1 & 78.4 & 92.0 \\ 
\midrule
\textbf{AEON (Ours)} & \textbf{75.5} & 79.8 & \textbf{94.1} \\
\midrule
\multicolumn{4}{l}{\textit{Estimated Noise Rates}} \\ 
\midrule
$\hat{\eta}^{id}$                             & 0.31 & \multicolumn{2}{c}{0.28} \\ 
$\hat{\eta}^{ood}$                            & 0.05 &  \multicolumn{2}{c}{0.17}  \\
\bottomrule
\end{tabular}
\end{table}

The fine-grained recognition results on WebFG-496~\cite{sun2021webly} presented in \cref{tab:web_fine_grained_comparison} further validate AEON's effectiveness across diverse visual domains. We achieve mean accuracy improvements $\Delta_{acc}$ of $2.29\%$, $3.19\%$, and $1.65\%$ on car, bird, and aircraft categories, respectively, compared to the strongest baseline~\cite{albert2023your}. Our approach yields consistent improvements on both synthetic and real-world corruptions while maintaining strong performance without accessing clean validation data. The consistent performance improvements highlight the method's ability to handle subtle noise patterns, showcasing its robustness.

An interesting observation from the real-world performance analysis presented in \cref{tab:clothing1m_webvision} is that the ranking of results in our synthetic benchmark (\cref{tab:places_idn_results}) closely mirrors those from the real-world datasets Clothing1M~\cite{xiao2015learning} and mini-WebVision~\cite{li2020dividemix}. This contrasts with the rankings observed in previously proposed benchmarks (\cref{tab:classification_comparison}), which show a different trend. Specifically, when focusing on the models common to both \cref{tab:places_idn_results} and \cref{tab:clothing1m_webvision}, our method achieves the highest performance, followed by MDM~\cite{fooladgar2024manifold}, ELR~\cite{liu2020early}, and DM~\cite{li2020dividemix}. Conversely, the results in \cref{tab:classification_comparison} present an inconclusive ranking, with MDM~\cite{fooladgar2024manifold} outperforming our AEON method in one of the experiments. Therefore, our benchmark offers a more reliable framework for assessing the effectiveness of new methods for learning with ID and OOD instance-dependent noisy labels.

\begin{table}[t]
\centering
\caption{\textbf{Real-world WebFG-496 Benchmarks.} Classification accuracy (\%) on WebFG-496~\cite{sun2021webly} fine-grained recognition benchmark across aircraft \(\textit{N}=100\), bird \(\textit{N}=200\), and car \(\textit{N}=196\) categories, with \textbf{bold} indicating SOTA.}
\label{tab:web_fine_grained_comparison}
\begin{tabular}{@{}l@{\hspace{8pt}}c@{\hspace{8pt}}c@{\hspace{8pt}}c@{}}
\toprule
Method & Web-Aircraft & Web-bird & Web-car \\ \midrule
CE & 60.80 & 64.40 & 60.60 \\
Co-teaching & 79.54 & 76.68 & 84.95 \\
PENCIL & 78.82 & 75.09 & 81.68 \\
SELFIE & 79.27 & 77.20 & 82.90 \\
DivideMix & 82.48 & 74.40 & 84.27 \\
Peer-learning & 78.64 & 75.37 & 82.48 \\
PLC & 79.24 & 76.22 & 81.87 \\ 
PLS & 87.58 & 79.00 & 86.27 \\ 
\midrule
\textbf{AEON (Ours)} & \textbf{89.23}  & \textbf{82.19} & \textbf{88.56} \\
\midrule
\multicolumn{4}{@{}l}{\textit{Estimated Noise Rates}} \\ 
\midrule
$\hat{\eta}^{id}$ & 0.21 & 0.18 & 0.28 \\
$\hat{\eta}^{ood}$ & 0.16 & 0.11 & 0.19 \\
\bottomrule
\end{tabular}
\end{table}

\begin{figure*}[t!]
   \centering
   \includegraphics[width=0.9\linewidth]{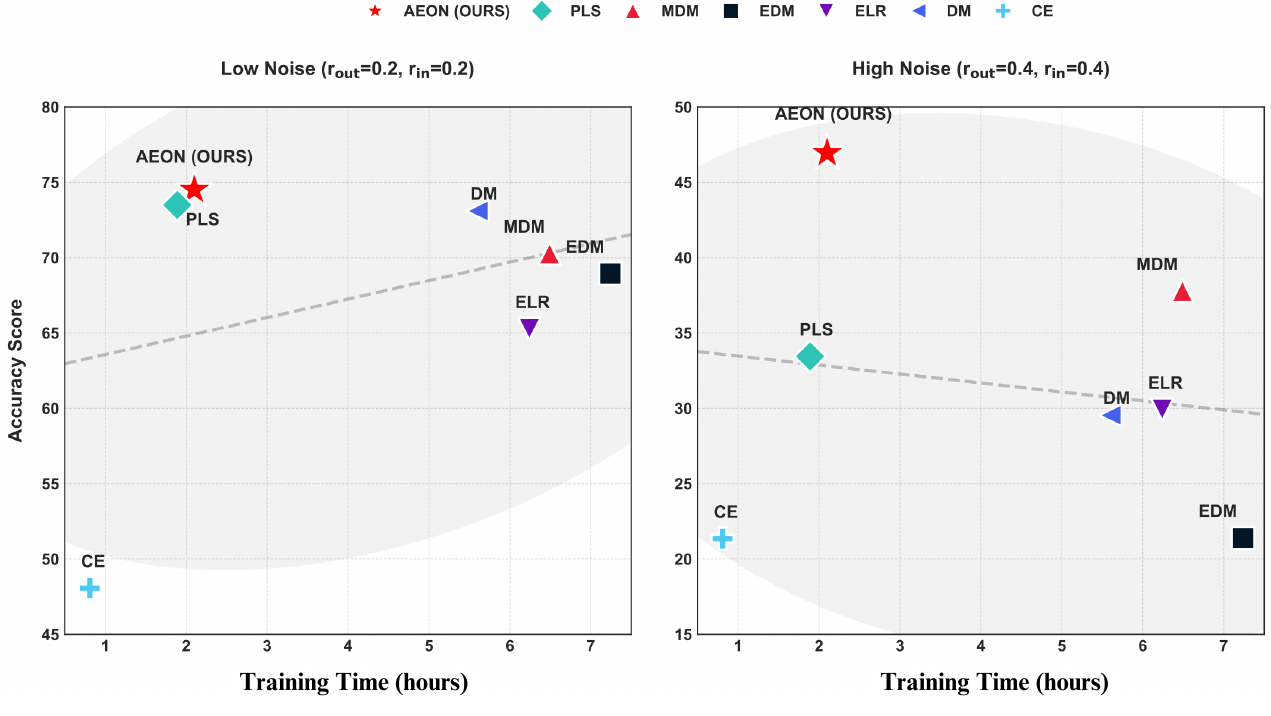}
   \caption{
   Performance analysis comparing training time efficiency and classification accuracy across methods. Results on our ciFAIR-100~\cite{barz2020we} benchmarks under low-noise $(r^{ood}=0.2, r^{id}=0.2)$ and high-noise $(r^{ood}=0.4, r^{id}=0.4)$ settings with instance-dependent closed-set noise~\cite{xia2020part} and instance-dependent Places-IDN~\cite{zhou2017places} open-set noise. Scatter plots illustrate the accuracy-training time trade-off, with \(95\%\) confidence regions shown in gray.}
   \label{fig:model-performance-comparison}
\end{figure*}

\section{Ablation Studies}
\label{sec:ablation}

We conduct comprehensive ablation studies on our proposed ciFAIR-100~\cite{barz2020we} benchmark with \(40\%\) closed-set~\cite{xia2020part} and \(40\%\) open-set noise Places365~\cite{zhou2017places}. Our analysis focuses on three key aspects: (1) comparative model performance under varying noise conditions in~\cref{subsec:performance_analysis}, (2) component contributions in~\cref{subsec:component_ablation}, and (3) initialization robustness in ~\cref{subsec:ablation_initialisation}. All experiments maintain consistent architecture and optimization settings as described in \cref{subsec:implementations}.

\subsection{Performance Analysis}
\label{subsec:performance_analysis}

\cref{fig:model-performance-comparison} presents a comprehensive performance analysis of various models under both low and high noise conditions. In the low noise scenario ($r^{ood}=0.2$, $r^{id}=0.2$), AEON and PLS~\cite{albert2023your} achieve superior accuracy ($74.51\% \pm 0.32$ and $73.50\% \pm 0.28$) with efficient training times ($2.10$ and $1.89$ hours). The high noise scenario ($r^{ood}=0.4$, $r^{id}=0.4$) demonstrates AEON's robust accuracy ($46.94\% \pm 0.41$) compared to other methods, with PLS following at $33.45\% \pm 0.37$, while maintaining computational efficiency. Models like EDM~\cite{sachdeva2021evidentialmix} and CE show significantly degraded performance under increased noise, despite having high computational requirements. The dashed trend lines in both plots and the gray confidence ellipses (\(95\%\) confidence region) reveal a consistent pattern: under low noise, accuracy exhibits positive correlation with computational time ($\rho=0.42$), while under high noise, this trend reverses ($\rho=-0.38$), suggesting that increased computation time does not guarantee better performance in challenging noise conditions.

\begin{table*}[t]
\caption{\textbf{Component-wise ablation study showing the progressive development of AEON}. Results on our  ciFAIR-100~\cite{barz2020we} benchmark with $40\%$ IDN~\cite{xia2020part} closed-set $r^{id}$, and $40\%$ open-set noise $r^{ood}$ using Places-IDN~\cite{zhou2017places} demonstrate how each component addresses specific aspects of noisy label learning. The model configurations show cumulative addition of components, with the noise estimation column showing AEON's estimated noise rates $\hat{\eta}^{id}$, $\hat{\eta}^{ood}$, and last column shows the overall training time in hours (h).}
\label{tab:component_ablation}
\centering
\scalebox{0.9}{
\begin{tabular}{@{}lccccc@{}}
\toprule
\multirow{2}{*}{\textbf{Model Configuration}} & \textbf{Accuracy (\%)} & \multicolumn{2}{c}{\textbf{Noise Estimation}} & \textbf{Train time} \\
\cmidrule(lr){3-4} 
& $\uparrow$ & $\hat{\eta}^{id}$ & $\hat{\eta}^{ood}$ & (h) \\
\midrule
CE Only (Base) [\cref{eq:supervised_loss}] & 21.35 & -- & -- & 0.81 \\
Closed-set: Sup/Unsup Learning [\cref{eq:supervised_loss,eq:unsup}] & 29.38 & -- & -- & 1.20 \\
+ Open-set: Energy Scoring [\cref{eq:energy}] & 31.67 & -- & -- & 1.35 \\
+ Masking: Adaptive Rate Est. [\cref{eq:w_dist,eq:w_clean,eq:eta_out,eq:eta_in}] & 39.20 & 0.41 & 0.38 & 1.89 \\
+ Cont: Contrastive Learning [\cref{eq:cont}] & \textbf{46.94} & 0.44 & 0.36 & 2.10\\
\bottomrule
\end{tabular}}
\end{table*}

\subsection{Component Analysis}
\label{subsec:component_ablation}

In \cref{tab:component_ablation}, the baseline cross-entropy (CE) model \cref{eq:supervised_loss}, achieves only $21.35\%$ accuracy, highlighting the severity of mixed closed-set and open-set noise in our benchmark. We first address closed-set noise through a combination of supervised learning \cref{eq:supervised_loss}, and unsupervised learning \cref{eq:unsup}, which improves accuracy to $29.38\%$ by better handling in-distribution label corruptions. This improvement is achieved with a modest increase in computational cost from $0.81$ to $1.20$ hours.

Building upon this foundation, we incorporate energy-based scoring \cref{eq:energy} to target open-set noise precisely. This addition further improves accuracy to $31.67\%$, demonstrating the complementary benefits of handling both noise types. The computational overhead remains relatively stable at $1.35$ hours.

A significant accuracy improvement is achieved with our dual-stream sigmoid soft masking mechanism \cref{eq:w_dist,eq:w_clean}, which adaptively weights sample contributions through $w^{ood}$ and $w^{id}$, raising accuracy to $39.20\%$. More significantly, this approach enables dynamic estimation of noise rates through learnable parameters \cref{eq:eta_in,eq:eta_out}, yielding estimates of $\hat{\eta}^{id} = 0.41$ and $\hat{\eta}^{ood} = 0.38$. Finally, integrating contrastive learning from \cref{eq:cont} yields our complete model, achieving $46.94\%$ accuracy through enhanced feature discrimination. The estimated noise rates ($\hat{\eta}^{id} = 0.44$, $\hat{\eta}^{ood} = 0.36$) validate our adaptive estimation approach.

Such improvement in accuracy is achieved with a training time of $2.10$ hours, representing a  $2.6\times$ increase over the baseline (CE). This gradual improvement demonstrates each component's efficacy and incremental integration in AEON.

\subsubsection{Sensitivity to Temperature Parameters}\label{subsubsec:temp_parameters}
We analyze AEON's sensitivity to three key temperature parameters: energy scoring $T_E$ in \cref{eq:energy}, noise rate estimation $T^{ood} = T^{id} = T$ in \cref{eq:eta_out,eq:eta_in}, and sigmoid masking $\beta^{old} = \beta^{id} = \beta$ in \cref{eq:w_dist,eq:w_clean}. As shown in \cref{tab:temp_sensitivity}, low  values for $T_E$ (e.g., $0.5$) lead to poor noise discrimination, resulting in low accuracy of $41.23\%$, while high values (e.g., $2.0$) tend to over-smooth the energy distribution, producing an accuracy of  $40.81\%$. The noise rate temperature $T$ affects estimation stability, with extreme values of $5.0$ or  $15.0$ showing suboptimal performance. Similarly, the sigmoid temperature $\beta$ balances transition sharpness, where $\beta=0.05$ is too smooth and $\beta=0.2$ is too sharp. The optimal configuration $T_E=1.0$, $T=10.0$, and $\beta=0.1$ achieves $46.94\%$ accuracy with noise estimates of $\hat{\eta}^{id}=0.44$ and $\hat{\eta}^{ood}=0.36$.

\begin{table}[t]
\caption{\textbf{Impact of temperature parameters on AEON's performance}. Results on our benchmark, ciFAIR-100~\cite{barz2020we} with $40\%$ closed-set IDN~\cite{xia2020part} $r^{id}$ and $40\%$ open-set noise $r^{ood}$ using Places-IDN~\cite{zhou2017places}, show AEON's noise estimated rates $\hat{\eta}^{id}$, $\hat{\eta}^{ood}$ and test accuracy.
For clarity, we maintain $T^{ood} = T^{id} = T$ and $\beta^{ood} = \beta^{id} = \beta$ throughout all experiments.  Configuration with highest accuracy is highlighted in \textbf{bold}.}
\label{tab:temp_sensitivity}
\small
\begin{tabular}{@{}ccccccc@{}}
\toprule
\textbf{$T_E$} & \textbf{$T$} & \textbf{$\beta$} & \multicolumn{2}{c}{\textbf{Noise Est.}} & \textbf{Acc.}  \\
\cmidrule(lr){4-5}
& & & $\hat{\eta}^{id}$ & $\hat{\eta}^{ood}$ & (\%) $\uparrow$  \\
\midrule
0.5 & 10.0 & 0.1 & 0.38 & 0.42 & 41.23 \\
2.0 & 10.0 & 0.1 & 0.35 & 0.45 & 40.81  \\
\midrule
1.0 & 5.0 & 0.1 & 0.41 & 0.39 & 43.67  \\
1.0 & 15.0 & 0.1 & 0.42 & 0.38 & 44.12 \\
\midrule
1.0 & 10.0 & 0.05 & 0.40 & 0.40 & 43.89  \\
1.0 & 10.0 & 0.2 & 0.43 & 0.37 & 45.03 \\
\midrule
\textbf{1.0} & \textbf{10.0} & \textbf{0.1} & 0.44 & 0.36 & \textbf{46.94}  \\
\bottomrule
\end{tabular}
\end{table}

\subsubsection{Architecture Analysis}\label{subsubsec:arch_analysis}
Following standard protocols in noise-robust learning literature~\cite{li2020dividemix,albert2023your,garg2024noiserate}, we implement AEON using PreActResNet-18~\cite{he2016identity} as our backbone architecture. To demonstrate AEON's architectural generality, we conduct a controlled study examining different backbone networks on ciFAIR-100~\cite{barz2020we} in \cref{tab:arch_analysis}. Results show that improved accuracy results can be achieved with more sophisticated models, such as ViT-Small~\cite{dosovitskiy2020image} and Swin-T~\cite{liu2021swin}. However, the noise estimation is relatively stable for all different models.

\begin{table}[t]
\caption{\textbf{Impact of different architectures on AEON's performance}. Results on our benchmark,  ciFAIR-100~\cite{barz2020we} with $40\%$ closed-set IDN~\cite{xia2020part} $r^{id}$ and $40\%$ open-set noise $r^{ood}$ using Places-IDN~\cite{zhou2017places}, show accuracy and estimated noise rates ($\hat{\eta}^{id}$ and  $\hat{\eta}^{ood}$) as a function of the different architectures.}
\label{tab:arch_analysis}
\small
\begin{tabular}{@{}lccc@{}}
\toprule
\textbf{Architecture} & \textbf{Acc.} & \multicolumn{2}{c}{\textbf{Est. Noise Rates}} \\
& (\%) $\uparrow$ & $\hat{\eta}^{id}$ & $\hat{\eta}^{ood}$ \\
\midrule
PreActResNet-18~\cite{he2016identity} & 46.94 & 0.36 & 0.44 \\
ResNet-50~\cite{he2016deep} & 48.62 & 0.38 & 0.42 \\
ConvNeXt-T~\cite{liu2022convnet} & 51.83 & 0.34 & 0.45 \\
ViT-Small~\cite{dosovitskiy2020image} & 52.41 & 0.35 & 0.46 \\
Swin-T~\cite{liu2021swin} & 52.15 & 0.33 & 0.45 \\
\bottomrule
\end{tabular}
\end{table}

\subsection{Initialization Sensitivity}\label{subsec:ablation_initialisation}
We analyze the sensitivity of AEON to different initializations of the learnable parameters $\gamma^{ood}$ and $\gamma^{id}$ that control our sigmoid-based noise rate estimation \cref{eq:eta_out,eq:eta_in}. \cref{tab:init_sensitivity} demonstrates the model's robustness across various initialisation scales on ciFAIR-100~\cite{barz2020we} with our instance-dependent noise injection protocol described in \cref{subsec:benchmark_explained}.

The results reveal a relatively low sensitivity of our method with respect to the initialization scale, with little variation of accuracy and estimated noise rates. 
Notably, random initialization using $\gamma \sim \mathcal{U}[-1,1]$ achieves comparable performance to fixed-scale initialization, with noise rate estimates within $\pm0.02$ of the optimal values. The stable performance consistency across initialization validates our sigmoid-based noise rate estimation approach.

\begin{table}[t]
\caption{\textbf{Analysis of initialization sensitivity for noise rate estimation parameters $\gamma^{ood} = \gamma^{id} = \gamma$}. Results on our benchmark, ciFAIR-100~\cite{barz2020we} with $40\%$ closed-set IDN~\cite{xia2020part} $r^{id}$, and $40\%$ open-set noise $r^{ood}$ using Places-IDN~\cite{zhou2017places}, show AEON's accuracy and estimated noise rate ($\hat{\eta}^{id}$, $\hat{\eta}^{ood}$) as a function of the initialization scale $|\gamma|$. }
\label{tab:init_sensitivity}
\begin{tabular}{@{}lccc@{}}
\toprule
\textbf{Init. Scale} & \textbf{Acc.} & \multicolumn{2}{c}{\textbf{Est. Noise Rates}} \\
$|\gamma|$ & (\%) $\uparrow$ & $\hat{\eta}^{id}$ & $\hat{\eta}^{ood}$ \\
\midrule
$10^{-2}$ & 46.21 & 0.34 & 0.42 \\
$10^{-1}$ & 46.58 & 0.35 & 0.43 \\
$1.0$     & 46.81 & 0.36 & 0.48 \\
$10.0$    & 46.52 & 0.36 & 0.42 \\
$\mathcal{U}[-1,1]$ & 46.75 & 0.34 & 0.45 \\
\bottomrule
\end{tabular}
\end{table}

\section{Discussion}\label{sec:discussion}
AEON advances ID and OOD instance-dependent label noise handling through dynamic estimation of noise rates ($\hat{\eta}^{clean}$, $\hat{\eta}^{ood}$), achieving SOTA performance across benchmarks. The framework demonstrates positive societal impact by mitigating dataset biases through effective noise detection. While manual tuning of temperature parameters $T_E$, $T$, and $\beta$ controlling noise estimation and sample weighting remains a limitation, AEON's robust performance across architectures and initialization strategies suggests broad applicability.

Furthermore, our synthetic benchmark results (\cref{tab:places_idn_results}) align closely with the rankings observed in real-world datasets like Clothing1M and mini-WebVision (\cref{tab:clothing1m_webvision}), where our method outperforms others. However, previously proposed benchmarks (\cref{tab:classification_comparison}) show inconclusive rankings, with MDM surpassing our method in one experiment. Hence, our benchmark provides a more effective approach for evaluating the performance of new noisy-label learning methods that handle ID and OOD instance-dependent label noise.

Future work should explore automatic temperature adaptation, establish theoretical bounds for noise rate estimation, and evaluate scalability to larger datasets with thousands of classes once they are available.

\section{Conclusion}\label{sec:conclusion}
AEON represents a significant advancement in handling noisy labels for image classification, being the first method to jointly estimate both in-distribution (ID) and out-of-distribution (OOD) instance-dependent noise rates ($\hat{\eta}^{id}$, $\hat{\eta}^{ood}$) without requiring clean validation data or hard thresholds. Through its novel sigmoid-based soft masking mechanism \cref{eq:w_dist,eq:w_clean}, AEON achieves SOTA performance across multiple benchmarks with minimal computational overhead (our training is $2.6\times$ slower than a baseline based on CE training and $1.2\times$ slower than previous SOTA). Additionally, we introduce a novel instance-dependent synthetic benchmark that better represents real-world noise, providing the community with a more realistic testbed for evaluating noise-robust methods. On our proposed benchmark and challenging real-world datasets like Clothing1M and WebFG-496, AEON's ability to accurately estimate noise rates while maintaining high classification accuracy demonstrates its practical applicability. This work opens new avenues for research in adaptive noise estimation techniques and their integration with semi-supervised learning frameworks. The success of AEON's soft masking approach suggests that future work exploring dynamic, instance-dependent noise handling mechanisms could further advance learning with noisy labels.

\bibliography{sn-bibliography}
\bibliographystyle{plain}  

\end{document}